\documentclass{article}
\PassOptionsToPackage{numbers}{natbib}
\usepackage[preprint]{neurips_2026}
\usepackage[utf8]{inputenc}
\usepackage[T1]{fontenc}
\usepackage{hyperref}
\usepackage{url}
\usepackage{graphicx}
\usepackage{booktabs}
\usepackage{amsmath}
\usepackage{amsfonts}
\usepackage{nicefrac}
\usepackage{microtype}
\usepackage{xcolor}
\usepackage{colortbl}
\usepackage{capt-of}
\usepackage{multirow}
\hypersetup{
  colorlinks=false,
  pdfborder={0 0 1},
  linkbordercolor={1 0 0},
  citebordercolor={0 1 0},
  urlbordercolor={0 0 1},
  filebordercolor={0 0 1},
  pdftitle={SPHQuant: Efficient extreme low bit weight quantization for Vision-Language Models},
  pdfauthor={Kewei Zhang, Zheng Chen, Haotong Qin, Yulun Zhang}
}
\title{SPHQuant: Efficient extreme low bit weight quantization for Vision-Language Models}
\author{
  Kewei Zhang$^{1}$\quad Zheng Chen$^{1}$\quad Haotong Qin$^{2}$\quad Yulun Zhang$^{1,}$\thanks{Corresponding author.}\\[0.4em]
  {\normalfont $^{1}$Shanghai Jiao Tong University}\\
  {\normalfont $^{2}$The Hong Kong Polytechnic University}
}

\begin{document}
\maketitle
\setcounter{footnote}{0}
\raggedbottom

\begin{abstract}
Recent foundation models are moving toward native multimodal Vision-Language Models (VLMs), making VLMs a central form of next-generation foundation models. However, their large language backbones make edge deployment difficult due to high memory footprint and memory-bound autoregressive decoding. Weight-only post-training quantization is a practical solution, but pushing VLMs to extreme low bit-widths remains challenging: existing rotation-free methods suffer from outliers at 2--3 bits, while rotation-based methods improve accuracy at the cost of additional runtime overhead. We propose SPHQuant, a rotation-free spherical weight-only quantization framework for VLMs. Instead of quantizing weights directly in Cartesian coordinates, SPHQuant decomposes each 8D weight vector into coordinate signs, radius, and a positive unit direction. This representation isolates outlier magnitude into the radius while keeping directions bounded and statistically regular. Based on this insight, SPHQuant allocates extra precision to the radius to mitigate accuracy degradation induced by outliers. It further uses a compact positive-direction codebook and fine-tunes codebook entries through angular parameterization to preserve the unit-sphere constraint. We also design a hardware-friendly GEMV kernel that keeps the direction codebook small enough for shared-memory lookup and packs radial bits efficiently. Experiments show that SPHQuant matches the performance of state-of-the-art extreme low-bit quantization methods while improving decode throughput over QTIP by \textbf{30.3\%} on RTX A6000.
Code will be released in \url{https://github.com/Pushazf/SPHQuant}.
\end{abstract}

\section{Introduction}
\label{sec:introduction}

Recent foundation models are moving beyond text-only language models toward unified vision-language foundation models~\cite{jin2025efficientmllmsurvey,qwen3.5,bai2025qwen3vltechnicalreport}. Rather than treating vision modules as add-ons to text-only LLMs, recent models jointly learn from vision and text tokens during training, making multimodal understanding a native capability. This shift makes VLMs an important model family and creates a growing need for edge deployment. Many VLM applications, such as mobile assistants~\cite{chu2023mobilevlm,chu2024mobilevlmv2}, on-device OCR, document understanding, embodied agents, and real-time visual reasoning, require low latency, privacy preservation, and reduced dependence on network connectivity. However, deploying VLMs on edge devices remains challenging: their billion-scale parameters require large device memory, while memory-bound autoregressive inference limits decode throughput.

\begin{figure}[t]
  \centering
  \includegraphics[width=\linewidth]{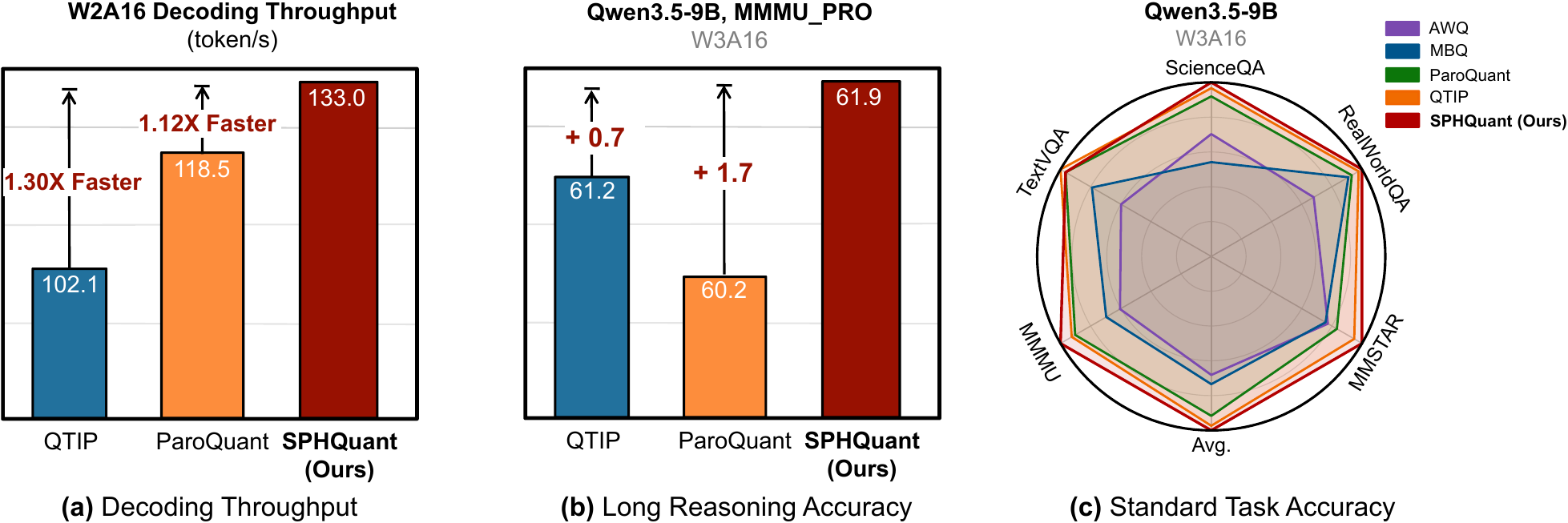}
  \caption{Performance of SPHQuant. (a) W2A16 decoding throughput. (b) Long-reasoning accuracy. (c) W3A16 task accuracy, normalized by the per-task maximum; radial range: 0.60--1.00.}
  \label{fig:main}
\end{figure}

Quantization has been explored across visual tasks, including image demoir\'eing~\cite{chen2025quantdemoire} and video super-resolution~\cite{chai2026quantvsr,wu2026lsgquant}. For VLM deployment, weight-only post-training quantization (PTQ) provides a practical way to reduce weight storage and memory traffic during autoregressive decoding. Although 4-bit weight-only quantization has been widely adopted~\cite{lin2023awq}, pushing VLMs to more aggressive 2 or 3 bit settings remains necessary for fitting larger multimodal models on memory-constrained devices. Existing methods, however, still struggle to balance accuracy and efficiency at such extreme bit-widths. QTIP achieves state-of-the-art 2 bit weight-only quantization with randomized-Hadamard incoherence processing~\cite{tseng2024qtip}, but its transforms introduce additional inference operations; ParoQuant reduces rotation overhead through pairwise Givens rotations~\cite{liang2026paroquant}, but its accuracy rapidly degrades at 2--3 bits due to the limitations of linear quantization. Preserving outlier weights in FP16 can improve accuracy but requires hardware-unfriendly sparse-dense computation.

This motivates a question: \textbf{can we handle weight outliers with low overhead under extreme low-bit VLM quantization?} We argue that the difficulty is not only caused by the limited bit-width, but also by the coordinate system in which weights are represented and quantized.

Most existing weight-only quantization methods operate directly in Cartesian coordinates. In this representation, a single weight outlier can expand the quantization range of the entire group, reducing the effective information density assigned to non-outlier weights. This effect is tolerable at moderate bit-widths but especially damaging at 2 or 3 bits.

\begin{figure}[!b]
  \centering
  \includegraphics[width=\linewidth]{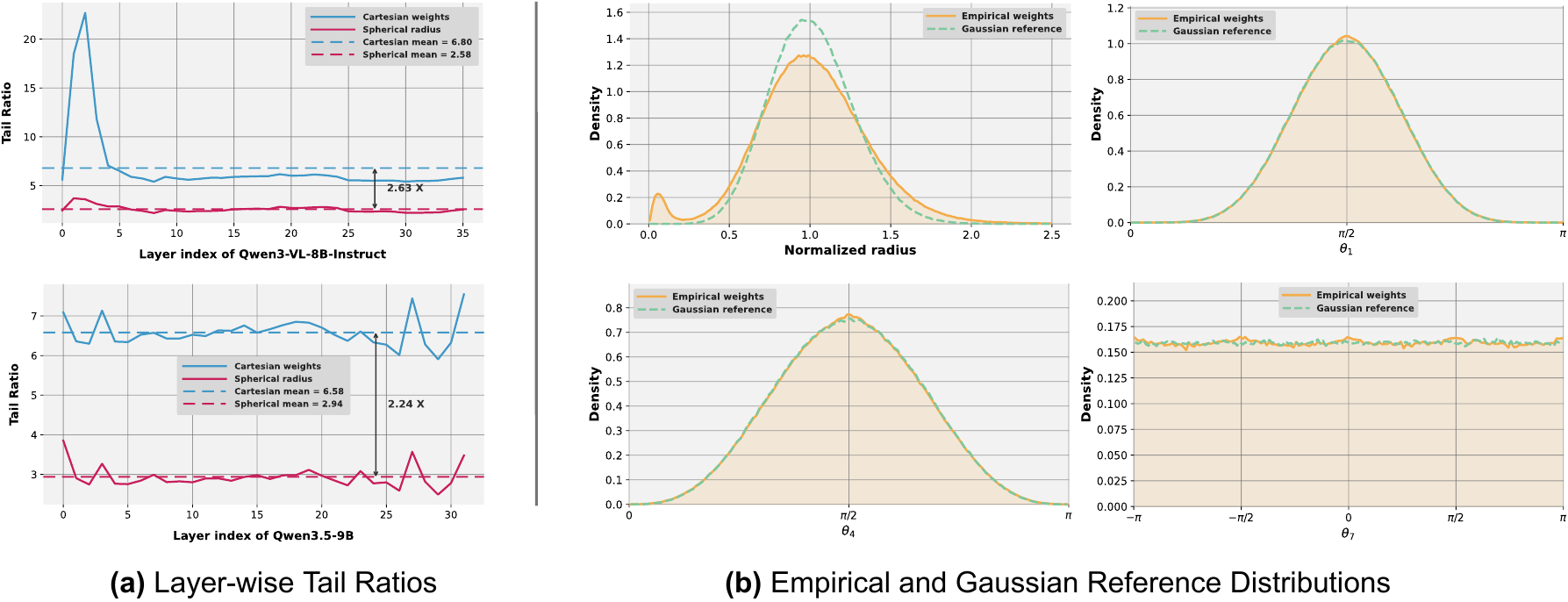}
  \caption{Observations that support the advantages of quantizing weights in spherical coordinates. (a) Layer-wise Tail Ratios. Spherical radius has significantly smaller tail ratios than Cartesian weights across all layers. (b) Empirical and Gaussian Reference Distributions. The weights exhibit an angular distribution that closely matches the Gaussian reference distribution, while their radial distribution deviates substantially.}
  \label{fig:observation}
\end{figure}

Instead of quantizing weights directly in Cartesian space, we propose to represent each 8D weight vector in spherical coordinates, where the radius and direction are explicitly decoupled. Given a weight vector $\mathbf{w}=\{w_i \mid i=1,\ldots,k\}\in\mathbb{R}^k$ with $k=8$, we transform it into its spherical representation $\mathbf{w}'=\{\phi_1,\phi_2,\ldots,\phi_{k-1},r\}$, where

\begin{equation}
r=\|\mathbf{w}\|_2, \quad \phi_i=\arccos\left(\dfrac{w_i}{\sqrt{\sum_{j=i}^{k}w_j^2}}\right), \quad i=1,\ldots,k-2, \quad \phi_{k-1}=\mathrm{atan2}(w_k,w_{k-1}).
\label{eq:spherical-coordinates}
\end{equation}
In this representation, the large values caused by outliers are captured by the radial component, while the directional component is constrained to a unit-sphere manifold. Two empirical observations motivate quantization in spherical coordinates and lead to a radial bit-allocation strategy. \textbf{First,} the radius is smoother than the original weights. For example, $(1,1,1,1,1,1,1,1)$ is transformed into $(2\sqrt{2}, \theta_1,\ldots,\theta_7)$, while $(1,1,1,1,5,1,1,1)$ is transformed into $(4\sqrt{2}, \theta'_1,\ldots,\theta'_7)$; the coordinate range increases from 1 to 5, but the radius only doubles. We quantify the tail ratio of weights in Cartesian coordinates and radii in spherical coordinates by dividing the 99.9th-percentile value by the 50th-percentile value. As shown in Fig.~\ref{fig:observation}a, across different layers, the radius distribution is consistently smoother than the raw weight distribution, making spherical quantization more robust to outliers. \textbf{Second,} even without rotation, the direction is already close to uniform. We decompose each 8D weight vector into seven angles and compare their distributions with the angle distributions of random Gaussian vectors, which correspond to uniformly sampled unit directions. As shown in Fig.~\ref{fig:observation}b, the angle of empirical directions closely match that of uniform-sphere reference, suggesting that directions of weights are close to uniform, and a direction codebook is sufficient for direction quantization. \textbf{Consequently,} since outlier magnitudes are isolated in the radius, we can allocate more bits to the radial component. For example, under 2-bit quantization, assigning 8 sign bits, a 6-bit positive-direction index, and 4 radius bits to each 8D vector adds 12.5\% to the encoded weight payload while further reducing the impact of outliers.

Based on these insights, we propose SPHQuant, a rotation-free spherical weight-only quantization framework for VLMs. We further design an efficient codebook fine-tuning method and a hardware-friendly CUDA kernel. Specifically, direct fine-tuning of 8D unit vectors in direction codebook requires projecting updates back to the unit sphere, which can discard useful update components and reduce optimization efficiency. We therefore parameterize direction codebook entries by angles to preserve the unit-sphere constraint by construction, and densely pack extra radial bits across 8D vectors for regular CUDA loads and fused lookup, dequantization, and matrix multiplication.

As shown in Fig.~\ref{fig:main}, SPHQuant achieves competitive accuracy on multiple recent VLMs under extreme low-bit weight-only quantization, while improving decode throughput over QTIP by 30.3\% on RTX A6000. In summary, our contributions include:
\begin{itemize}
  \item We identify a rotation-free spherical representation for extreme low-bit VLM quantization, isolating outlier magnitude in the radius while keeping directions close to uniform.
  \item We propose SPHQuant, which decouples each 8D weight vector into radius and direction, compactly quantizes directions, and allocates extra bits to the radius for outlier robustness.
  \item We design angle-parameterized codebook fine-tuning and a CUDA-friendly kernel with dense radial-bit packing and fused dequantization for efficient inference.
  \item We demonstrate competitive extreme low-bit accuracy across recent VLMs and a 30.3\% decoding-throughput improvement over QTIP on RTX A6000.
\end{itemize}

\section{Related Work}

\paragraph{Vision-language models.} Modern LLM-based VLMs combine visual perception with language backbones for multimodal understanding and reasoning~\cite{wang2025internvl3_5,qwen36_35b_a3b}. As VLMs evolve from adapter-based architectures toward native multimodal foundation models such as Qwen3.5~\cite{qwen3.5},  Gemma4~\cite{farabet2026gemma4}, they enable applications including mobile assistants, OCR, document understanding, embodied agents, and real-time visual reasoning. However, their language backbones dominate parameter storage and decode-stage memory traffic, making weight-only PTQ important for deployment. Extreme 2- and 3-bit VLM quantization remains challenging because it must preserve multimodal accuracy while achieving real speedups with efficient low-bit kernels.

\paragraph{Weight-only post-training quantization.}
Weight-only PTQ reduces model memory and decode traffic without full retraining. The simplest baseline is round-to-nearest (RTN) linear quantization, which maps a weight group to $b$-bit integer codes with a uniform scale and zero point:
\begin{equation}
Q(\mathbf{w})=\operatorname{clip}\left(\left\lfloor\frac{\mathbf{w}}{s}\right\rceil+z,0,2^b-1\right),\quad
s=\frac{\max(\mathbf{w})-\min(\mathbf{w})}{2^b-1},\quad
z=-\left\lfloor\frac{\min(\mathbf{w})}{s}\right\rceil .
\label{eq:rtn}
\end{equation}
Scalar methods such as GPTQ~\cite{frantar-gptq} and AWQ~\cite{lin2023awq} are hardware-friendly at 4 bits, but become fragile near 2 bits due to outliers and limitation of linear quantization. Transform and vector-code methods improve ultra-low-bit accuracy: QuIP/QuIP\# use incoherence processing~\cite{chee2024quip2bitquantizationlarge,tseng2024quipbetterllmquantization}, QTIP uses trellis-coded quantization~\cite{tseng2024qtip}, and ParoQuant reduces rotation overhead through pairwise rotations~\cite{liang2026paroquant}. Direction--magnitude decomposition has also been explored in PVQ~\cite{vanderouderaa2024pyramid} and PCDVQ~\cite{yue2025pcdvqenhancingvectorquantization}. PVQ combines Hadamard coherence processing with an implicit integer-lattice representation of directions and separate gain quantization. PCDVQ uses randomized Hadamard regularization and distribution-aligned direction and magnitude codebooks. SPHQuant instead quantizes the original vectors without rotation, adapts its direction codebook through angular fine-tuning, and allocates extra precision to the radius to handle large magnitudes. Our contribution lies in this combination of representation, bit allocation, optimization, and efficient decoding rather than the direction--magnitude decomposition alone. VLM-specific methods such as MBQ~\cite{li2024mbq} and VEQ~\cite{qin2026veq} model modality sensitivity or expert heterogeneity, but they mainly improve calibration objectives under conventional linear quantization.

\section{Method}
\label{headings}

In this section, we introduce SPHQuant, a novel weight-only quantization framework for VLMs. We first describe how SPHQuant performs spherical weight quantization and fine-tunes the quantization parameters. Then, we explain how to implement an efficient CUDA kernel for SPHQuant.

\subsection{Quantization in Spherical Coordinates}
\label{sec:spherical-quantization}

\begin{figure}[t]
  \centering
  \includegraphics[width=\linewidth]{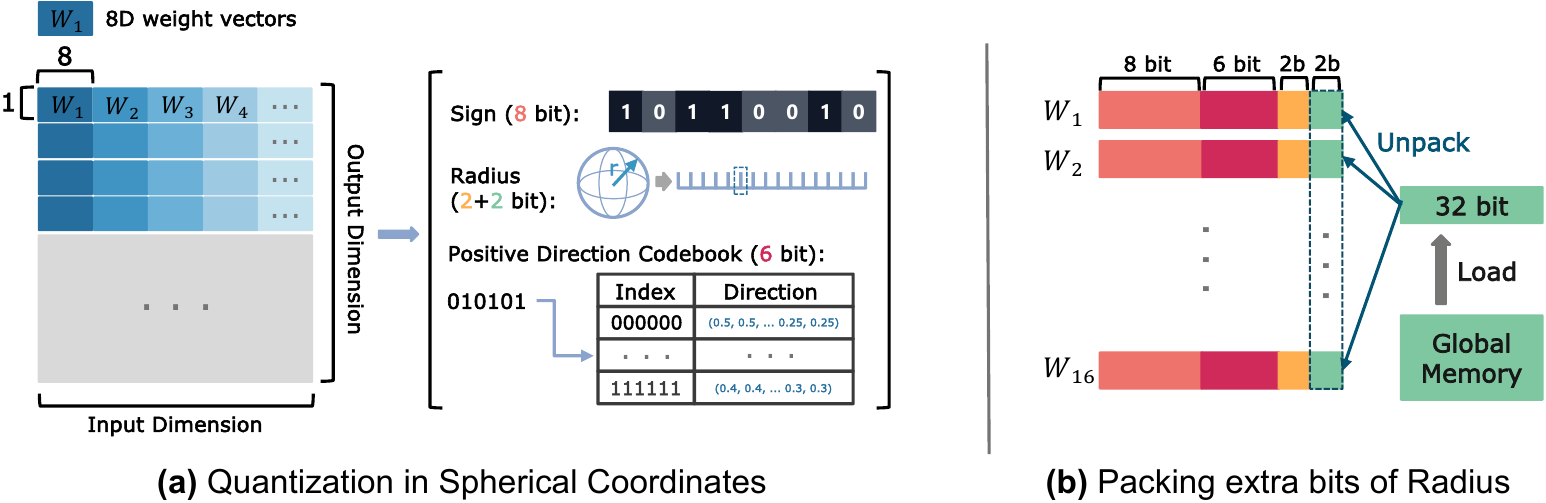}
  \caption{Overview of SPHQuant.
(a) In \textbf{W2A16} setting, each 8D weight vector is quantized in spherical coordinates by decomposing it into signs, radius, and direction. The signs are stored exactly with 8 bits, the radius is linearly quantized with 4 bits including 2 extra bits, and the direction is quantized using a codebook of positive unit vectors.
(b) Packed extra radius bits. The extra 2 radius bits from every 16 8D weight vectors are packed and loaded as one 32-bit word.
}
  \label{fig:overview}
\end{figure}

SPHQuant applies spherical quantization to the weights of selected linear layers in the VLM language backbone. Given a weight matrix $W\in\mathbb{R}^{d_{\mathrm{out}}\times d_{\mathrm{in}}}$, we split each row along the input dimension into contiguous 8D vectors $\mathbf{w}\in\mathbb{R}^{8}$. Each vector is decomposed into coordinate signs, radius, and nonnegative unit direction:
\begin{equation}
\mathbf{s}=\operatorname{sign}(\mathbf{w}), \quad \mathbf{w}_{+}=|\mathbf{w}|,\quad r=\|\mathbf{w}_{+}\|_2,\quad \mathbf{d}=\frac{\mathbf{w}_{+}}{\max(r,\epsilon)}.
\label{eq:sph-decompose}
\end{equation}
Here $\mathbf{w}_{+}$ denotes the positive-orthant version of $\mathbf{w}$, $\mathbf{s}\in\{-1,+1\}^{8}$, $r\ge 0$, and $\mathbf{d}\in\mathbb{S}^{7}_{+}$, where $\mathbb{S}^{7}_{+}=\{\mathbf{u}\in\mathbb{R}^{8}_{\ge 0}:\|\mathbf{u}\|_2=1\}$ is the positive orthant of the 8D unit sphere. Reconstruction gives

\begin{equation}
\hat{\mathbf{w}}=\hat r\cdot \mathbf{s}\odot \hat{\mathbf{d}}.
\label{eq:sph-reconstruct}
\end{equation}
where $\hat r$ is the dequantized radius, $\hat{\mathbf{d}}$ is the dequantized positive unit direction, and $\mathbf{s}$ is the sign vector extracted during quantization. For each 8D vector, $\mathbf{s}$ can be stored exactly with 8 sign bits and remains unchanged during quantization. We next describe radius and positive-direction quantization and dequantization, and the efficiency benefit of storing signs separately.

For the positive unit direction, SPHQuant uses a codebook $\mathcal{C}=\{\mathbf{c}_k\}$ with $\mathbf{c}_k\in\mathbb{S}^{7}_{+}$. Let $B$ be the target weight bit-width. Since the spherical direction has seven degrees of freedom, we allocate $7B$ bits to direction-related information. The coordinate signs are stored separately with 8 bits per 8D vector, leaving a $K_{\mathrm{dir}}=7B-8$ bit index for the positive-direction codebook. Thus $|\mathcal{C}|=2^{K_{\mathrm{dir}}}=2^{7B-8}$. Separating signs is important for efficient lookup: with sign separation, the codebook only needs to cover positive unit directions, whereas a signed-direction codebook would need an additional factor of $2^8$ entries to represent all coordinate sign patterns. In the 2-bit setting, i.e., $B=2$, the direction codebook has $2^6$ entries. Since each entry stores an 8D unit vector in fp16, the sign-separated codebook requires 1 KiB, while the signed-direction codebook requires $2^8$ times more storage, or 256 KiB. Since modern GPUs provide only a few hundreds of KiB of shared memory per SM, and prior work also emphasizes avoiding large lookup tables~\cite{tseng2024qtip}, sign separation makes the direction codebook small enough for shared-memory lookup in the 2-bit setting.

Because the directions are empirically close to uniformly distributed on the unit sphere, spherical k-means can be viewed as a data-adaptive refinement of a uniform spherical codebook: it preserves broad directional coverage while adapting the centers to the actual layer weights. We sample normalized directions with probability proportional to $r_i^2$, so high-magnitude vectors have stronger influence on the codebook, and assign each direction index by cosine similarity:

\begin{equation}
z_i=\arg\max_{k\in\{1,\ldots,2^{K_{\mathrm{dir}}}\}} \mathbf{d}_i^\top \mathbf{c}_k.
\label{eq:sph-assign}
\end{equation}
Here $z_i$ is the direction index assigned to the $i$-th 8D weight vector, and $\hat{\mathbf{d}}_i=\mathbf{c}_{z_i}$ is its dequantized positive unit direction. Since both $\mathbf{d}_i$ and $\mathbf{c}_k$ are unit vectors, maximizing their inner product is equivalent to choosing the codebook direction with the smallest angular distance.
Given a fixed direction index, SPHQuant does not directly quantize the original radius $r_i$. Instead, it first fits a pre-quantization radius $\tilde r_i$ along the assigned direction:
\begin{equation}
\tilde r_i=r_i^{\mathrm{LS}}=\arg\min_{\alpha\ge 0}\left\|(\mathbf{w}_i)_{+}-\alpha\mathbf{c}_{z_i}\right\|_2^2=\langle (\mathbf{w}_i)_{+},\mathbf{c}_{z_i}\rangle,
\label{eq:sph-ls-radius}
\end{equation}
where $\langle\cdot,\cdot\rangle$ is the Euclidean inner product; the last equality holds because $\mathbf{c}_{z_i}$ is a unit vector. Thus $\tilde r_i$ is the input to radius quantization, while $\hat r_i$ is its dequantized value for reconstruction.

SPHQuant stores $\tilde r_i$ with group-wise uniform quantization. Radius groups are formed along the input dimension of each weight row: each group contains 128 consecutive scalar weights, or equivalently 16 consecutive 8D weight vectors. Let $\mathcal{G}$ denote one such group, and let $i\in\mathcal{G}$ index an 8D weight vector in the group. Suppose each radius is allocated $V$ bits. We initialize $\tilde r_{\min}^{\mathcal{G}}=\min_{i\in\mathcal{G}}\tilde r_i$, $\tilde r_{\max}^{\mathcal{G}}=\max_{i\in\mathcal{G}}\tilde r_i$, and $\Delta_{\mathcal{G}}=(\tilde r_{\max}^{\mathcal{G}}-\tilde r_{\min}^{\mathcal{G}})/(2^V-1)$. We quantize and dequantize each radius as
\begin{equation}
q_i=\operatorname{clip}\left(\left\lfloor\frac{\tilde r_i-\tilde r_{\min}^{\mathcal{G}}}{\Delta_{\mathcal{G}}}\right\rceil,0,2^V-1\right),\quad \hat r_i=\tilde r_{\min}^{\mathcal{G}}+q_i\Delta_{\mathcal{G}},\quad i\in\mathcal{G}.
\label{eq:sph-radius-quant}
\end{equation}
The $i$-th vector is reconstructed as $\hat{\mathbf{w}}_i=\hat r_i\cdot\mathbf{s}_i\odot\mathbf{c}_{z_i}$.

\subsection{Outlier-Aware Radial Bit Allocation}
\label{sec:radial-bit-allocation}

As discussed in Sec.~\ref{sec:introduction}, spherical coordinates isolate the effect of weight outliers into the radius. Compared with quantizing weights in Cartesian coordinates, spherical quantization concentrates outliers in the radius only, so we can mitigate the outlier problem by allocating extra bits to the radius. To verify that the accuracy gain from increasing the radius bit-width comes from mitigating outliers rather than simply using more bits, we introduce a matched-bit control experiment on layer 0 of Qwen3-VL-8B-Instruct.

\noindent
\begin{minipage}[t]{0.49\linewidth}
  \vspace{0pt}
  For SPHQuant, we quantize the layer with the 2-bit direction setting in Sec.~\ref{sec:spherical-quantization}.

  We vary the radius bit-width $V\in\{2,3,4\}$, and measure the output reconstruction loss against the full-precision layer after the fine-tuning procedure introduced in Sec.~\ref{sec:angular-finetuning}.
We report the relative loss reduction using each method's own 2-bit result as the zero point.
\end{minipage}
\hfill
\begin{minipage}[t]{0.47\linewidth}
  \vspace{0pt}
  \centering
  \setlength{\abovecaptionskip}{2pt}
  \setlength{\belowcaptionskip}{8pt}
  \includegraphics[width=\linewidth]{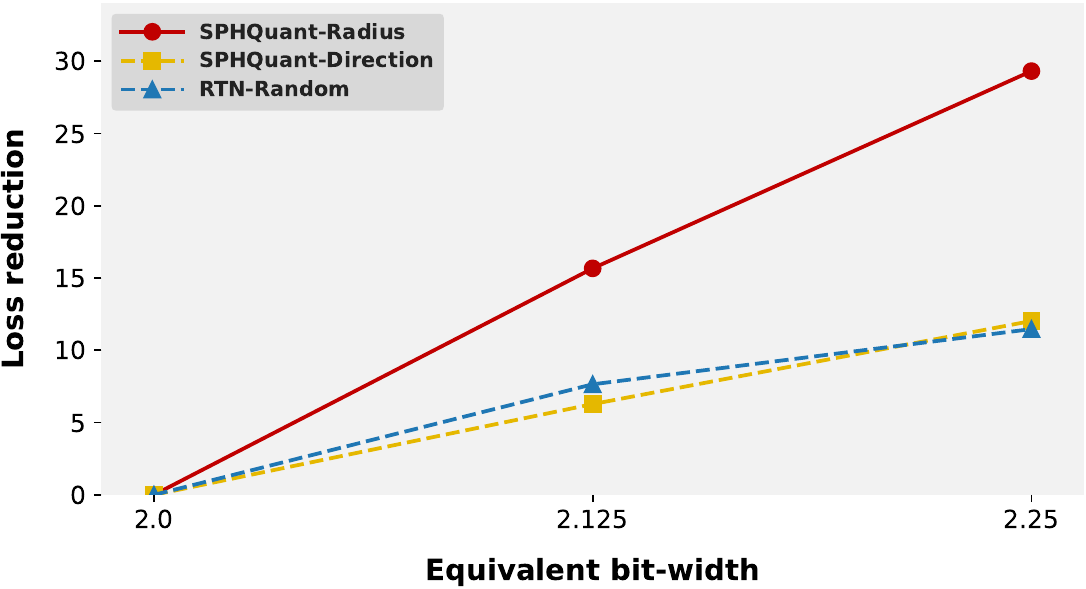}
  \captionof{figure}{Loss reduction under matched average bit-widths. Higher is better.}
  \label{fig:radial-bit-allocation}
  \vspace{0.4em}
\end{minipage}

As a Cartesian baseline, we use RTN quantization and randomly assign higher precision to $1/8$ of the weights. The three matched settings are: all weights use 2-bit RTN for 2.00 equivalent bits; randomly selected $1/8$ weights use 3 bits and the rest use 2 bits for 2.125 equivalent bits; randomly selected $1/8$ weights use 4 bits and the rest use 2 bits for 2.25 equivalent bits. These match SPHQuant's average bit-width $2+(V-2)/8$ for $V=2,3,4$, respectively.

From Fig.~\ref{fig:radial-bit-allocation}, as the equivalent bit-width increases from 2.00 to 2.25, SPHQuant improves much faster than the random mixed-precision RTN baseline. This indicates that the gain is not merely due to using more bits, but comes from allocating the extra precision to the radial component where outlier magnitude is concentrated. We further discuss how allocating extra bits to the radius or direction affects final accuracy in Sec.~\ref{sec:ablation-study}.

\subsection{Efficient Layer-Wise Optimization with Angular Codebooks}
\label{sec:angular-finetuning}

After the spherical quantization step, the trainable parameters are the continuous radius parameters $\boldsymbol{\rho}$ and the direction codebook parameters $\boldsymbol{\phi}$. The codebook is initialized by radius-weighted spherical k-means, while the pre-quantization radii are initialized by projection onto the assigned directions.

Inspired by the layer-wise reconstruction strategy in ParoQuant~\cite{liang2026paroquant}, SPHQuant performs layer-wise reconstruction with two activation streams for each decoder layer. The full-precision stream provides the teacher output, and the quantized stream provides the student input produced by the already quantized preceding layers. The optimization objective is
\begin{equation}
\min_{\boldsymbol{\rho},\boldsymbol{\phi}}\mathbb{E}_{\mathbf{x}_q}\left[\left\|F_l^{\mathrm{SPH}}(\mathbf{x}_q;\boldsymbol{\rho},\boldsymbol{\phi})-F_l^{\mathrm{FP}}(\mathbf{x}_{\mathrm{fp}})\right\|_2^2\right],
\label{eq:sph-layer-recon}
\end{equation}
where $F_l^{\mathrm{SPH}}$ denotes the $l$-th decoder layer with selected linear modules replaced by SPHQuant pseudo-quantized modules, and $F_l^{\mathrm{FP}}$ denotes the original full-precision layer. Here $\mathbf{x}_{\mathrm{fp}}$ is the input activation to the full-precision layer, while $\mathbf{x}_q$ is the activation produced by the already quantized preceding layers. Optimizing the quantized layer on $\mathbf{x}_q$ allows later layers to adapt to errors introduced by earlier quantized layers, reducing error propagation in the full model. For the radius, we map each continuous parameter to a positive value by $r=\operatorname{softplus}(\rho)$ and apply $V$-bit group-wise fake quantization with a straight-through estimator during training.

\begin{figure}[t]

  \centering
  \includegraphics[width=0.72\linewidth]{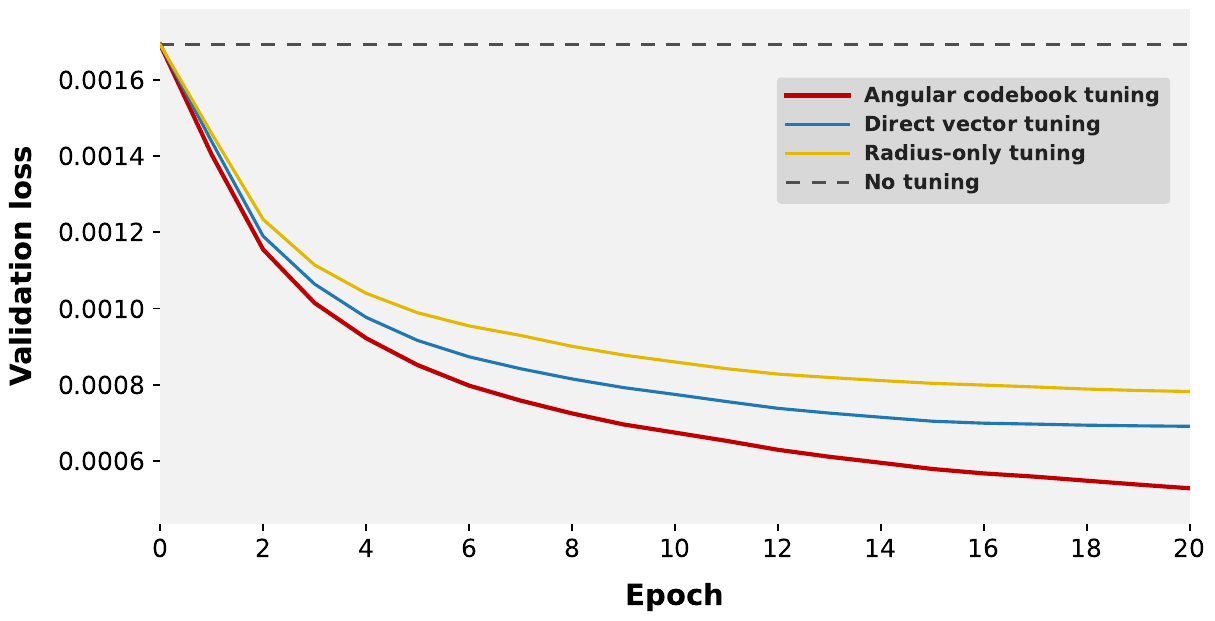}
  \caption{Loss curve of layer 0 of Qwen3-VL-8B-Instruct under different tuning strategies.}
  \label{fig:angular-codebook-tuning}

\end{figure}
\textbf{Angular codebook tuning.} The key design is to fine-tune directions through angles rather than directly updating 8D unit vectors. A Euclidean gradient on a unit codebook vector can contain components normal to the sphere. To keep the vector on the unit sphere, direct vector tuning must project the gradient onto the sphere or renormalize the vector after each update, which removes part of the update and reduces the effective gradient. SPHQuant avoids this inefficiency by parameterizing each codebook vector with seven hyperspherical angles:
\begin{equation}
[\mathbf{c}]_1=\cos\theta_1,\quad [\mathbf{c}]_i=\left(\prod_{j=1}^{i-1}\sin\theta_j\right)\cos\theta_i\ (2\le i\le 7),\quad [\mathbf{c}]_8=\prod_{j=1}^{7}\sin\theta_j.
\label{eq:sph-angle-codebook}
\end{equation}
To keep the optimization unconstrained while ensuring nonnegative unit directions, we store angle logits $\boldsymbol{\phi}\in\mathbb{R}^{7}$ and map them to
\begin{equation}
\theta_j=\epsilon+\left(\frac{\pi}{2}-2\epsilon\right)\sigma(\phi_j).
\label{eq:sph-angle-logits}
\end{equation}
Thus every forward pass reconstructs a valid codebook $\mathcal{C}(\boldsymbol{\phi})\subset\mathbb{S}^{7}_{+}$ by construction. The codebook indices $z_i$ are fixed after initialization; fine-tuning only moves the angular codebook and radii. As illustrated in Fig.~\ref{fig:angular-codebook-tuning}, angular codebook tuning reduces reconstruction loss more effectively in this experiment than radius-only tuning or direct vector tuning.

\subsection{Efficient Kernel Implementation}
\label{sec:efficient-kernel}

We implement efficient dequantization in a fused CUDA GEMV kernel~\cite{nvidia_cuda_programming_guide}. The following packed layout and 64-entry codebook apply to W2 with $B=2$ and $V=4$.

\textbf{Packed extra radius bits.}
Along the input dimension, every 128 weights form one quantization group, corresponding to 16 8D weight vectors. For each 8D vector, SPHQuant stores its packed quantized code in one 16-bit word: the lower 8 bits encode signs, the next 6 bits encode the direction codebook index, and the upper 2 bits encode the low bits of the radius code. Since the radius receives extra precision, each 128-weight group additionally stores one 32-bit word that packs the extra 2 radius bits of all 16 vectors. This word-aligned layout enables coalesced global-memory access.

\textbf{Shared-memory direction lookup.}
We preload the 64-entry direction codebook into shared memory and store each 8D direction as four \texttt{half2} pairs. During GEMV, each lane loads the direction by codebook index, restores coordinate signs from the packed sign bits, and multiplies the signed direction by the dequantized radius.

\textbf{Fused dequantization and accumulation.}
For each 8D vector, SPHQuant reconstructs the weight block only in registers. The reconstructed block is immediately multiplied with the fp16 activation values and accumulated in fp32. By fusing radius dequantization, direction lookup, sign restoration, and GEMV accumulation in one kernel, SPHQuant avoids writing dequantized weights to global memory while preserving the accuracy benefit of extra radial precision.

Sec.~\ref{app:sphquant-procedure} summarizes the full SPHQuant layer-wise quantization procedure, and Sec.~\ref{app:cuda-details} provides additional CUDA implementation details.

\section{Experiments}
\label{experiments}

\subsection{Experimental Settings}
\label{sec:experimental-settings}

\textbf{Models and Tasks.}
We evaluate SPHQuant on recent VLMs covering different model scales and architectures: Qwen3-VL-8B-Instruct~\cite{bai2025qwen3vltechnicalreport}, Qwen3.5-9B~\cite{qwen3.5}, and Gemma4-26B-A4B-it~\cite{google2026gemma4modelcard}. For standard multimodal evaluation, we use benchmarks covering college-level multimodal understanding: MMMU\_val~\cite{yue2024mmmumassivemultidisciplinemultimodal}; scene text understanding: TextVQA~\cite{singh2019vqamodelsread}; science question answering: ScienceQA~\cite{lu2022learnexplainmultimodalreasoning}; real-world visual understanding: RealWorldQA~\cite{xai2024realworldqa}; and general multimodal perception: MMStar~\cite{chen2024rightwayevaluatinglarge}. For reasoning-oriented evaluation, we enable the thinking mode of Qwen3.5-9B on MMMU-Pro~\cite{yue2025mmmuprorobustmultidisciplinemultimodal} to assess long-form multimodal reasoning accuracy.

\textbf{Baselines.}
We compare SPHQuant with two categories of weight-only PTQ baselines. Linear quantization methods: AWQ~\cite{lin2023awq}, MBQ~\cite{li2024mbq}, and ParoQuant~\cite{liang2026paroquant}. AWQ uses activation-aware scaling to protect salient weights, MBQ improves VLM quantization with modality-balanced calibration, and ParoQuant applies pairwise rotations to reduce rotation overhead. Vector quantization methods: QTIP~\cite{tseng2024qtip}, which uses trellis-coded quantization with incoherence processing to improve extreme low-bit accuracy. Since the novel architecture of Gemma4-26B-A4B depends on a specialized training procedure, we do not include ParoQuant or QTIP results on this model.

\textbf{Implementation Details.}
We use 2048 COCO samples~\cite{lin2015microsoftcococommonobjects} for calibration and reserve 128 samples for validation. Quantization is performed with a group size of 128. The direction codebook is initialized by spherical k-means with 65,536 sampled vectors and 20 iterations. We optimize the spherical quantization parameters with layer-wise reconstruction using MSE loss, where the student layer takes activations from the quantized stream. Joint optimization is run for 20 epochs with Adam~\cite{kingma2017adammethodstochasticoptimization}, using learning rates of $1\times10^{-3}$ for both radius parameters and codebook parameters. We select the best checkpoint according to validation loss. Layer-wise calibration is conducted on a single RTX PRO 6000 GPU with 96GB memory. For W2 evaluations, we use $V=4$ radius bits, yielding 2.25 encoded bits per weight; the Qwen3-VL-8B-Instruct W3 results in Tab.~\ref{tab:sota-results} use 3.125. Both payloads exclude group-wise parameters and direction codebooks; Sec.~\ref{sec:more-experiment-settings} details the storage accounting. For Qwen3-VL-8B-Instruct, calibration also preserves DeepStack visual features to match the real multimodal inference path; details are provided in Sec.~\ref{app:deepstack-calibration}.

\begin{table}
  \centering
  \caption{Comparison on standard VLM benchmarks. Bit denotes nominal quantization settings. SPHQuant W2 uses 2.25 encoded bits per weight; its Qwen3-VL-8B-Instruct W3 row uses 3.125. These payloads exclude group parameters and codebooks. See Tab.~\ref{tab:matched-bits} for matched-bit comparisons.}
  \label{tab:sota-results}
  \scriptsize
  \setlength{\tabcolsep}{3.2pt}
  \renewcommand{\arraystretch}{1.08}
  \resizebox{\linewidth}{!}{
  \begin{tabular}{l|l|l|ccccc|c}
    \toprule
    \rowcolor{green!10}
    & & & \multicolumn{5}{c|}{Tasks} & \\
    \rowcolor{green!10}
    \multicolumn{1}{c|}{\multirow{-2}{*}{Model}} & \multicolumn{1}{c|}{\multirow{-2}{*}{Bit}} & \multicolumn{1}{c|}{\multirow{-2}{*}{Method}} & MMMU\_val & TextVQA & ScienceQA & RealWorldQA & MMSTAR & \multicolumn{1}{c}{\multirow{-2}{*}{Avg.}} \\
    \midrule
    \multirow{9}{*}{\shortstack{Qwen3-VL-8B\\-Instruct}}
      & BF16 & BF16 & 52.9 & 82.4 & 93.1 & 69.7 & 68.0 & 73.2 \\
      \cmidrule{2-9}
      & \multirow{2}{*}{W2} & QTIP & 42.2 & \textbf{77.2} & 80.4 & 57.1 & \textbf{57.6} & 62.9 \\
      & & \cellcolor{gray!8}SPHQuant (Ours) & \cellcolor{gray!8}\textbf{42.2} & \cellcolor{gray!8}76.8 & \cellcolor{gray!8}\textbf{81.0} & \cellcolor{gray!8}\textbf{57.4} & \cellcolor{gray!8}57.5 & \cellcolor{gray!8}\textbf{63.0} \\
      \cmidrule{2-9}
      & \multirow{6}{*}{W3} & RTN & 46.9 & 75.4 & 84.2 & 63.1 & 60.8 & 66.1 \\
      & & AWQ & 49.8 & 78.9 & 87.3 & 65.9 & 63.7 & 69.1 \\
      & & MBQ & 49.3 & 79.8 & 88.4 & 65.2 & 63.7 & 69.3 \\
      & & ParoQuant & 50.5 & 79.5 & 88.2 & 68.3 & 64.6 & 70.2 \\
      & & QTIP & \textbf{52.1} & 80.5 & \textbf{92.0} & \textbf{70.7} & 66.8 & 72.4 \\
      & & \cellcolor{gray!8}SPHQuant (Ours) & \cellcolor{gray!8}51.7 & \cellcolor{gray!8}\textbf{81.0} & \cellcolor{gray!8}91.8 & \cellcolor{gray!8}69.9 & \cellcolor{gray!8}\textbf{67.6} & \cellcolor{gray!8}\textbf{72.4} \\
    \midrule
    \multirow{9}{*}{Qwen3.5-9B}
      & BF16 & BF16 & 47.4 & 69.1 & 92.5 & 67.1 & 64.8 & 68.2 \\
      \cmidrule{2-9}
      & \multirow{2}{*}{W2} & QTIP & \textbf{41.7} & 63.0 & \textbf{84.7} & \textbf{61.5} & 61.1 & \textbf{62.4} \\
      & & \cellcolor{gray!8}SPHQuant (Ours) & \cellcolor{gray!8}41.4 & \cellcolor{gray!8}\textbf{63.7} & \cellcolor{gray!8}83.0 & \cellcolor{gray!8}61.3 & \cellcolor{gray!8}\textbf{61.7} & \cellcolor{gray!8}62.2 \\
      \cmidrule{2-9}
      & \multirow{6}{*}{W3} & RTN & 38.3 & 44.5 & 78.8 & 56.3 & 50.5 & 53.7 \\
      & & AWQ & 36.8 & 56.4 & 79.3 & 57.8 & 57.3 & 57.5 \\
      & & MBQ & 38.4 & 61.6 & 73.5 & 63.9 & 56.9 & 58.9 \\
      & & ParoQuant & 42.0 & 66.3 & 87.1 & 64.5 & 58.8 & 63.7 \\
      & & QTIP & 42.4 & \textbf{67.2} & 88.8 & 65.7 & 61.7 & 65.2 \\
      & & \cellcolor{gray!8}SPHQuant (Ours) & \cellcolor{gray!8}\textbf{43.7} & \cellcolor{gray!8}66.3 & \cellcolor{gray!8}\textbf{90.0} & \cellcolor{gray!8}\textbf{66.3} & \cellcolor{gray!8}\textbf{63.0} & \cellcolor{gray!8}\textbf{65.9} \\
    \midrule
    \multirow{5}{*}{\shortstack{Gemma4-26B\\-A4B}}
      & BF16 & BF16 & 53.8 & 69.5 & 91.3 & 65.8 & 64.2 & 68.9 \\
      \cmidrule{2-9}
      & \multirow{4}{*}{W3} & RTN & 46.7 & 62.8 & 83.2 & 59.1 & 59.1 & 62.2 \\
      & & AWQ & 48.3 & 66.1 & 84.7 & 58.7 & 62.3 & 64.0 \\
      & & MBQ & 49.0 & 64.2 & 87.2 & 60.5 & \textbf{63.5} & 64.9 \\
      & & \cellcolor{gray!8}SPHQuant (Ours) & \cellcolor{gray!8}\textbf{51.4} & \cellcolor{gray!8}\textbf{67.0} & \cellcolor{gray!8}\textbf{88.7} & \cellcolor{gray!8}\textbf{64.6} & \cellcolor{gray!8}63.3 & \cellcolor{gray!8}\textbf{67.0} \\
    \bottomrule
  \end{tabular}
  }
\end{table}

\begin{table}
  \centering
  \begin{minipage}[t]{0.39\linewidth}
    \centering
    \caption{Accuracy of Qwen3.5-9B under long-reasoning generation.}
    \label{tab:reasoning-results}
    \scriptsize
    \setlength{\tabcolsep}{4.0pt}
    \renewcommand{\arraystretch}{1.08}
    \resizebox{\linewidth}{!}{
    \begin{tabular}{l|l|c}
      \toprule
      Bit & Method & MMMU-Pro \\
      \midrule
      BF16 & BF16 & 63.6 \\
      \cmidrule{1-3}
      \multirow{2}{*}{W2} & QTIP & {52.7} \\
      & \cellcolor{gray!8}SPHQuant (Ours) & \cellcolor{gray!8}\textbf{53.1} \\
      \cmidrule{1-3}
      \multirow{6}{*}{W3} & RTN & 43.1 \\
      & AWQ & 54.8 \\
      & MBQ & 57.1 \\
      & ParoQuant & 60.2 \\
      & QTIP & 61.2 \\
      & \cellcolor{gray!8}SPHQuant (Ours) & \cellcolor{gray!8}\textbf{61.9} \\
      \bottomrule
    \end{tabular}
    }
  \end{minipage}
  \hfill
  \begin{minipage}[t]{0.56\linewidth}
    \centering
    \caption{Decoding throughput (tokens/s) of Qwen3-VL-8B-Instruct on RTX A6000. SPHQuant W2 uses 2.25 encoded bits per weight.}
    \label{tab:throughput-results}
    \scriptsize
    \setlength{\tabcolsep}{4.0pt}
    \renewcommand{\arraystretch}{1.55}
    \resizebox{\linewidth}{!}{
    \begin{tabular}{l|l|cc}
      \toprule
      Bit & Method & Throughput & Speedup \\
      \midrule
      BF16 & BF16 & 43.45 & 1.00 \\
      \cmidrule{1-4}
      \multirow{4}{*}{W2} & AWQ & 140.57 & 3.24 \\
      & ParoQuant & 118.47 & 2.73 \\
      & QTIP & 102.05 & 2.35 \\
      & \cellcolor{gray!8}SPHQuant (Ours) & \cellcolor{gray!8}132.96 & \cellcolor{gray!8}3.06 \\
      \bottomrule
    \end{tabular}
    }
  \end{minipage}
\end{table}

\subsection{Comparison with State-of-the-Art Methods}
\label{sec:sota-comparison}

\textbf{Task Accuracy.}
We evaluate task accuracy on Qwen3-VL-8B-Instruct, Qwen3.5-9B, and Gemma 4-26B-A4B-it, on standard multimodal benchmarks with thinking disabled. We also evaluate the quantized Qwen3.5-9B model on MMMU-Pro with \texttt{enable\_thinking=True}. At 3 bits, we compare SPHQuant with all baselines. At 2 bits, since linear quantization methods largely collapse, we only compare SPHQuant with QTIP. As shown in Tab.~\ref{tab:sota-results}, SPHQuant outperforms linear quantization baselines and achieves accuracy comparable to the state-of-the-art extreme low-bit method QTIP.

\textbf{Decoding Throughput.}
We test decoding throughput of Qwen3-VL-8B-Instruct under the W2A16 setting on an RTX A6000 GPU. We compare AWQ, ParoQuant, QTIP, and SPHQuant at 2-bit weight quantization. For a fair comparison, all methods are evaluated with the same Transformers inference pipeline~\cite{wolf-etal-2020-transformers}, and we only replace the quantized linear modules with the corresponding method-specific CUDA kernels. The results in Tab.~\ref{tab:throughput-results} show that SPHQuant improves decoding throughput by 30.3\% over QTIP and 12.2\% over ParoQuant, while being only 5.4\% slower than AWQ. Sec.~\ref{sec:more-experiment-settings} discusses encoded bit-width and optimization cost.

Sec.~\ref{app:additional-experiments} reports additional baseline, matched-budget, and deployment evaluations.

\subsection{Ablation Study}
\label{sec:ablation-study}

Unless otherwise specified, all ablation experiments are conducted on Qwen3-VL-8B-Instruct under the W2A16 setting. We use 8D weight vectors, a group size of 128, 2048 COCO calibration samples, and 20 epochs of layer-wise fine-tuning.

\textbf{Ablation on Outlier-Aware Radial Bit Allocation.} Starting from the 2-bit SPHQuant setting, we compare two matched-budget allocation strategies: adding bits to the radial component, or using the same additional budget to enlarge the direction codebook. As shown in Tab.~\ref{tab:ablation-radial-bits}, allocating extra bits to the radius yields larger accuracy gains than assigning the same budget to direction indices. This supports our observation that outlier magnitude is mainly concentrated in the radius.

\textbf{Calibration cost of SPHQuant.} Tab.~\ref{tab:ablation-calib-epochs} reports sensitivity to calibration set size and fine-tuning epochs. Increasing the calibration set from 512 to 2048 samples at 20 epochs improves MMMU accuracy from 40.9 to 42.2, while 10 epochs with 2048 samples reach 42.0.

\begin{table}

  \centering
  \begin{minipage}[t]{0.52\linewidth}
    \centering
    \caption{Ablation on bit allocation.}
    \label{tab:ablation-radial-bits}
    \scriptsize
    \setlength{\tabcolsep}{1.5pt}
    \renewcommand{\arraystretch}{0.88}
    \resizebox{\linewidth}{!}{
    \begin{tabular}{l|ccc|cc|c}
      \toprule
      Variant & Sign & Dir. & Radius & Total & Eq. Bits & MMMU \\
      \midrule
      W2 SPHQuant & 8 & 6 & 2 & 16 & 2.000 & 37.2 \\
      +1 Radius Bit & 8 & 6 & 3 & 17 & 2.125 & 40.9 \\
      +1 Direction Bit & 8 & 7 & 2 & 17 & 2.125 & 38.8 \\
      +2 Radius Bits & 8 & 6 & 4 & 18 & 2.250 & \textbf{42.2}\\
      +2 Direction Bits & 8 & 8 & 2 & 18 & 2.250 & 39.5 \\
      \bottomrule
    \end{tabular}
    }
  \end{minipage}
  \hfill
  \begin{minipage}[t]{0.44\linewidth}
    \centering
    \caption{Calibration and epoch ablation.}
    \label{tab:ablation-calib-epochs}
    \scriptsize
    \setlength{\tabcolsep}{2.0pt}
    \renewcommand{\arraystretch}{0.78}
    \scalebox{1.15}{
    \begin{tabular}{cc|c}
      \toprule
      Calib. Samples & Epochs & MMMU \\
      \midrule
      512 & 20 & {40.9} \\
      1024 & 20 & {41.7} \\
      2048 & 5 & {41.2} \\
      2048 & 10 & 42.0 \\
      2048 & 20 & \textbf{42.2} \\
      \bottomrule
    \end{tabular}
    }
  \end{minipage}

\end{table}

\textbf{Effect of Angular Codebook Tuning.}
 We compare the initialization-only quantizer, radius-only tuning, and the full SPHQuant optimization that jointly tunes radius parameters and direction codebook angles. We also compare with direct vector tuning followed by renormalization. Tab.~\ref{tab:ablation-angular-tuning} shows that angular tuning improves over radius-only tuning, indicating that adapting the direction codebook on the unit sphere is important for improving tuning efficiency.

\begin{table}
  \centering
  \caption{Effect of angular codebook tuning.}
  \label{tab:ablation-angular-tuning}
  \scriptsize
  \setlength{\tabcolsep}{3.4pt}
  \renewcommand{\arraystretch}{1.08}
  \resizebox{\linewidth}{!}{
  \begin{tabular}{l|ccc|ccc}
    \toprule
    Variant & Radius Tune & Direction Tune & Parameterization & Valid Loss & MMMU & TextVQA \\
    \midrule
    No Tuning & No & No & Fixed codebook & 0.00169 & 23.4 & 62.9 \\
    Radius-only Tuning & Yes & No & Fixed codebook & 0.00078 & 39.7 & 73.8 \\
    Radius + Vector Tuning & Yes & Yes & Vector + renorm & 0.00069 & 40.5 & 74.2 \\
    Radius + Angular Tuning & Yes & Yes & Hyperspherical angles & \textbf{0.00052} & \textbf{42.2} & \textbf{76.8} \\
    \bottomrule
  \end{tabular}
  }
\end{table}

\textbf{Choice of Vector Dimension.}\phantomsection\label{sec:vector-dimension}
We compare the evaluated 4D, 8D, and 16D configurations on Qwen3-VL-8B-Instruct in Tab.~\ref{tab:vector-dimension}. Moving from 4D to 8D improves MMMU and TextVQA accuracy by 0.8 and 1.1 percentage points, respectively, while the 16D configuration adds only 0.2 points on each metric over 8D. The direction-codebook storage grows substantially: under the W2 direction allocation, the FP16 codebooks occupy 32 B, 1 KiB, and 512 KiB, respectively. The 16D codebook exceeds the shared-memory capacity of an RTX 4090 SM~\cite{nvidia_ada_tuning_guide}, while the 8D codebook is compact enough for our shared-memory lookup. These observations motivate 8D as a practical default for the current implementation. This does not establish optimality at a fixed storage budget.

\begin{table}[!htbp]
  \centering
  \caption{Accuracy and direction-codebook storage across vector dimensions on Qwen3-VL-8B-Instruct. Codebooks use FP16 directions under the W2 direction allocation.}
  \label{tab:vector-dimension}
  \small
  \setlength{\tabcolsep}{7pt}
  \begin{tabular}{cccc}
    \toprule
    Vector size & Direction codebook & MMMU\_val $\uparrow$ & TextVQA $\uparrow$ \\
    \midrule
    4D & 32 B & 41.4 & 75.7 \\
    8D & 1 KiB & 42.2 & 76.8 \\
    16D & 512 KiB & \textbf{42.4} & \textbf{77.0} \\
    \bottomrule
  \end{tabular}
\end{table}

\section{Conclusion}
In this paper, we propose SPHQuant, a spherical weight-only quantization framework for VLMs. By representing weights in spherical coordinates, SPHQuant decouples outlier magnitudes from bounded directional information and enables more effective bit allocation under extreme low-bit constraints. We further develop efficient optimization and hardware-friendly inference support to preserve accuracy while reducing runtime overhead. Experiments on recent VLMs show that SPHQuant achieves competitive extreme low-bit accuracy and improves decoding throughput by 30.3\% over QTIP on RTX A6000. SPHQuant combines outlier handling, codebook design, and kernel efficiency through a spherical representation for practical extreme low-bit VLM deployment.
\label{sec:conclusion}

\begingroup
\setlength{\bibsep}{0.5pt}

\endgroup

\clearpage
\section*{Appendix}

\section{SPHQuant Quantization Procedure}
\label{app:sphquant-procedure}
The following pseudocode summarizes the complete layer-wise quantization pipeline used by SPHQuant. The procedure follows the method described in Sec.~\ref{sec:spherical-quantization}--\ref{sec:angular-finetuning}: each selected linear layer is quantized in spherical coordinates, the radius and angular codebook parameters are optimized by layer-wise reconstruction, and the final representation is packed for efficient inference.

\begingroup
\small
\newcommand{\sphstep}[3]{\textcolor{black!55}{#1} & \begingroup\raggedright\leftskip=#2\relax #3\par\endgroup\\[2pt]}
\begin{center}
\begin{minipage}{\linewidth}
\hrule height 0.6pt
\vspace{4pt}
\noindent\textbf{Pseudocode: SPHQuant layer-wise quantization}\par
\vspace{4pt}
\hrule height 0.4pt
\vspace{5pt}
\noindent\textbf{Input:} full-precision VLM $M$, calibration set $\mathcal{D}$, selected linear layers, vector dimension $k=8$, target bit-width $B$, radius bit-width $V$, group size $G=128$, fine-tuning epochs $T$.\par
\vspace{3pt}
\noindent\textbf{Output:} quantized VLM: packed signs, direction indices, radius codes, group scales, and tuned codebooks.\par
\vspace{5pt}
\hrule height 0.4pt
\vspace{5pt}
\noindent\begin{tabular}{@{}r@{\hspace{0.7em}}p{\dimexpr\linewidth-2em\relax}@{}}
\sphstep{1}{0pt}{Run the full-precision model on calibration samples and collect the input activations of the first language layer. Split calibration samples into training and validation subsets.}
\sphstep{2}{0pt}{Initialize two activation streams: the full-precision stream $\mathbf{x}_{\mathrm{fp}}$ and the quantized stream $\mathbf{x}_q$.}
\addlinespace[3pt]
\sphstep{3}{0pt}{\textbf{For} each decoder layer $l$ \textbf{do}: compute the full-precision teacher output $F_l^{\mathrm{FP}}(\mathbf{x}_{\mathrm{fp}})$.}
\sphstep{4}{1.2em}{\textbf{For} each selected weight matrix $W$ in layer $l$ \textbf{do}: split $W$ row-wise into contiguous 8D vectors $\mathbf{w}_i$.}
\sphstep{5}{2.4em}{Decompose each vector into coordinate signs, radius, and positive unit direction:\newline
$\mathbf{s}_i=\operatorname{sign}(\mathbf{w}_i)$, $(\mathbf{w}_i)_+=|\mathbf{w}_i|$, $r_i=\|(\mathbf{w}_i)_+\|_2$, and $\mathbf{d}_i=(\mathbf{w}_i)_+/\max(r_i,\epsilon)$.}
\sphstep{6}{2.4em}{Initialize the positive-direction codebook by radius-weighted spherical k-means; assign $z_i$ by cosine similarity and fit pre-quantization radii by least-squares projection onto assigned directions.}
\sphstep{7}{2.4em}{Parameterize radii by continuous variables $\boldsymbol{\rho}$ with softplus mapping, and parameterize each codebook entry by angular variables $\boldsymbol{\phi}$ so that all directions remain on $\mathbb{S}^{7}_{+}$.}
\sphstep{}{1.2em}{\textbf{end for}}
\sphstep{8}{1.2em}{Replace selected linear modules in layer $l$ with SPHQuant pseudo-quantized modules. During each forward pass, apply group-wise $V$-bit radius fake quantization with a straight-through estimator and reconstruct weights as $\hat{\mathbf{w}}_i=\hat r_i\mathbf{s}_i\odot\mathbf{c}_{z_i}$.}
\sphstep{9}{1.2em}{Optimize $\boldsymbol{\rho}$ and $\boldsymbol{\phi}$ for $T$ epochs using layer-wise reconstruction loss between $F_l^{\mathrm{SPH}}(\mathbf{x}_q)$ and $F_l^{\mathrm{FP}}(\mathbf{x}_{\mathrm{fp}})$, and select the checkpoint with the lowest validation loss.}
\sphstep{10}{1.2em}{Finalize discrete radius codes, radius group minima and scales, sign bits, direction indices, and optimized direction codebook entries. For the W2 setting with $V=4$, pack each 8D vector into the 16-bit base code and pack the extra radius bits of every 16 vectors into one auxiliary 32-bit word.}
\sphstep{11}{1.2em}{Update activation streams: set $\mathbf{x}_q \leftarrow F_l^{\mathrm{SPH}}(\mathbf{x}_q)$ and $\mathbf{x}_{\mathrm{fp}} \leftarrow F_l^{\mathrm{FP}}(\mathbf{x}_{\mathrm{fp}})$, then proceed to the next layer.}
\sphstep{}{0pt}{\textbf{end for}}
\addlinespace[3pt]
\sphstep{12}{0pt}{\textbf{Return} the quantized model and packed CUDA-friendly metadata for inference.}
\end{tabular}
\vspace{3pt}
\hrule height 0.6pt
\end{minipage}
\end{center}
\endgroup

\section{Additional Experiment Settings}
\label{sec:more-experiment-settings}

\paragraph{Effective bit-width and throughput fairness.}
SPHQuant assigns two extra radial bits to each 8D vector in the W2 setting, resulting in an encoded weight payload of 2.25 bits per weight, excluding group-wise quantization parameters and codebook storage. This extra budget is used only for the radius, as motivated by Sec.~\ref{sec:radial-bit-allocation}, where outlier magnitude is concentrated. Although this gives SPHQuant 0.25 more bits than strict 2-bit methods, the comparison should be interpreted with this storage difference in mind. Additional bits increase memory traffic, while Tab.~\ref{tab:throughput-results} shows that SPHQuant is still faster than QTIP and ParoQuant. This is enabled by the compact positive-direction codebook and packed radial-bit layout described in Sec.~\ref{sec:efficient-kernel}.

\paragraph{Additional decoding evaluation.}
The original throughput experiment uses RTX A6000. The additional evaluation uses RTX 4090 with 24 GB of memory, Qwen3-VL-8B-Instruct, batch size 1, 256 prefill tokens, and 512 generated tokens. Tab.~\ref{tab:rtx4090-throughput} reports the additional throughput results, and Tab.~\ref{tab:gpu-memory} reports measured GPU-memory consumption. These quantities are presented separately from encoded bits per weight because they capture different aspects of deployment cost.

\paragraph{Matched encoded weight budgets.}
The additional comparison in Tab.~\ref{tab:matched-bits} evaluates RTN, ParoQuant, and SPHQuant at 3.00 and 3.125 encoded bits per weight on Qwen3-VL-8B-Instruct. For SPHQuant with 8D vectors, the encoded payload is $(7B+V)/8$ bits per weight; it excludes group-wise quantization parameters and direction codebooks. The SPHQuant W3 row for Qwen3-VL-8B-Instruct in Tab.~\ref{tab:sota-results} corresponds to the 3.125-bit result. Keeping these payloads explicit avoids interpreting nominal W2/W3 labels as complete model-storage measurements.

\paragraph{Additional training settings.}
For fair comparison, we keep the optimization budget of QTIP, ParoQuant, and SPHQuant in a similar single-GPU range. Since the three methods use different optimization procedures, e.g., QTIP optimizes linear layers individually, ParoQuant~\cite{liang2026paroquant} uses a two-stage optimization pipeline, and SPHQuant performs layer-wise spherical reconstruction, we match their practical quantization cost rather than forcing identical hyperparameters. Tab.~\ref{tab:more-training-settings} reports the training samples, epochs, wall-clock time, and device used for Qwen3-VL-8B-Instruct.

\begin{table}
  \centering
  \caption{Additional training settings on Qwen3-VL-8B-Instruct.}
  \label{tab:more-training-settings}
  \scriptsize
  \setlength{\tabcolsep}{5.0pt}
  \renewcommand{\arraystretch}{1.08}
  \begin{tabular}{l|ccc|c}
    \toprule
    Method & Training Samples & Epochs & Time & Device \\
    \midrule
    QTIP & 512 & 10 & 8h & 1$\times$ RTX PRO 6000 \\
    ParoQuant & 1024 & 10 & 6h & 1$\times$ RTX PRO 6000 \\
    SPHQuant & 2048 & 20 & 6h & 1$\times$ RTX PRO 6000 \\
    \bottomrule
  \end{tabular}
\end{table}

\section{DeepStack-Aware Calibration for Qwen3-VL}
\label{app:deepstack-calibration}

Qwen3-VL-8B-Instruct~\cite{bai2025qwen3vltechnicalreport} uses DeepStack features to inject visual information into selected language decoder layers. During calibration, SPHQuant explicitly preserves this multimodal path. The calibration inputs are first preprocessed by \texttt{process\_model}, and the decoder layers that should receive DeepStack features are recorded through \texttt{\_mbq\_calib\_deepstack\_layers}. This carries visual DeepStack features with each calibration batch to the corresponding decoder layers.

When SPHQuant captures the first-layer hidden states and the layer-wise keyword arguments, it also stores these DeepStack-related intermediate features as part of the layer input context. Therefore, during layer-wise reconstruction, each quantized layer receives the same type of multimodal context as in the original forward pass. This avoids text-only hidden-state approximation and aligns calibration with the actual vision-language inference path.

\section{Additional CUDA Implementation Details}
\label{app:cuda-details}

The following details describe the fused GEMV implementation for the W2 configuration with 8D vectors, a 64-entry positive-direction codebook, and four radius bits. They supplement the packed-weight layout in Sec.~\ref{sec:efficient-kernel}.

\paragraph{Thread mapping and activation access.}
Each thread block contains 256 threads and computes eight output channels. Each thread processes 32 FP16 activations, corresponding to four 8D weight vectors. This mapping exposes parallelism across output channels and input-vector groups while allowing the packed codes and activations to be consumed within the same kernel.

\paragraph{Shared-memory codebook.}
The $64\times8$ FP16 direction codebook occupies 1 KiB and is staged in shared memory. A transposed layout arranges codebook coordinates for lookup by the decoding threads and reduces access conflicts. Direction indices select the positive unit vectors, and the separately stored sign bits restore coordinate signs during reconstruction.

\paragraph{Packed decoding and fused computation.}
Bit operations extract the coordinate signs, six-bit direction index, and low radius bits from each 16-bit base word. The additional 32-bit word for a group supplies the extra radius bits of its 16 vectors. After assembling the radius code, a fused multiply-add applies the group's scale and offset to reconstruct the radius. Direction lookup, sign restoration, radius reconstruction, and multiplication with the activations are fused, avoiding a global-memory buffer of dequantized weights.

\paragraph{Vectorized products and reduction.}
The kernel uses vectorized \texttt{half2} instructions for pairwise operations on FP16 values and accumulates partial results in FP32. Warp-shuffle operations and shared-memory reductions combine thread-local contributions into the output channels. The CUDA kernel will be released with the implementation.

\section{Additional Comparisons and Deployment Experiments}
\label{app:additional-experiments}

\subsection{Comparison with Direction--Magnitude Quantization}
\label{app:direction-magnitude}
PVQ~\cite{vanderouderaa2024pyramid} and PCDVQ~\cite{yue2025pcdvqenhancingvectorquantization} provide closely related approaches that separately represent weight direction and magnitude. Tab.~\ref{tab:direction-magnitude} compares their evaluated implementations with SPHQuant on Qwen3-VL-8B-Instruct. SPHQuant improves MMMU accuracy by 2.0 and 2.4 percentage points over PVQ and PCDVQ, respectively, and TextVQA by 3.1 and 2.6 percentage points. It also achieves higher measured decoding throughput in this comparison. Together, these results support the practical effectiveness of SPHQuant's combination of rotation-free quantization, adaptive angular codebooks, and radial bit allocation; they do not isolate the contribution of any one component.

\begin{table}[!htbp]
  \centering
  \caption{Additional comparison with direction--magnitude quantizers on Qwen3-VL-8B-Instruct. SPHQuant uses its W2 configuration with 2.25 encoded bits per weight.}
  \label{tab:direction-magnitude}
  \small
  \setlength{\tabcolsep}{8pt}
  \begin{tabular}{lccc}
    \toprule
    Method & MMMU\_val $\uparrow$ & TextVQA $\uparrow$ & Tokens/s $\uparrow$ \\
    \midrule
    PVQ & 40.2 & 73.7 & 209.03 \\
    PCDVQ & 39.8 & 74.2 & 210.11 \\
    SPHQuant & \textbf{42.2} & \textbf{76.8} & \textbf{288.63} \\
    \bottomrule
  \end{tabular}
\end{table}

\subsection{Decoding on RTX 4090}
\label{app:rtx4090-throughput}
We further evaluate Qwen3-VL-8B-Instruct on an RTX 4090 with 24 GB of memory, using batch size 1, a prefill length of 256 tokens, and a decoding length of 512 tokens. As shown in Tab.~\ref{tab:rtx4090-throughput}, SPHQuant reaches 288.63 tokens/s, or 2.70$\times$ the FP16 throughput. It is 37.6\% faster than QTIP and 4.9\% faster than ParoQuant in this setting, while RTN remains slightly faster. This extends the throughput evaluation to a GPU with less memory than RTX A6000.

\begin{table}[!htbp]
  \centering
  \caption{Batch-one decoding throughput of Qwen3-VL-8B-Instruct on RTX 4090-24GB, with 256 prefill tokens and 512 generated tokens. Payload excludes auxiliary storage.}
  \label{tab:rtx4090-throughput}
  \small
  \setlength{\tabcolsep}{8pt}
  \begin{tabular}{lccc}
    \toprule
    Method & Payload (bits/weight) & Tokens/s $\uparrow$ & Speedup \\
    \midrule
    FP16 & 16 & 106.85 & 1.00$\times$ \\
    QTIP & 2.00 & 209.76 & 1.96$\times$ \\
    ParoQuant & 2.00 & 275.07 & 2.57$\times$ \\
    SPHQuant & 2.25 & 288.63 & 2.70$\times$ \\
    RTN & 2.00 & \textbf{300.059} & \textbf{2.81$\times$} \\
    \bottomrule
  \end{tabular}
\end{table}

\subsection{Measured GPU-memory Consumption}
\label{app:gpu-memory}
Tab.~\ref{tab:gpu-memory} reports measured GPU-memory consumption for the quantized Qwen3-VL-8B-Instruct model under the same inference configuration across the three methods. SPHQuant uses 5.35 GB, compared with 5.02 GB for QTIP and 5.24 GB for ParoQuant. Thus, its measured footprint is 0.33 GB and 0.11 GB larger, respectively. Encoded weight payload and GPU-memory consumption are distinct quantities: the latter also includes unquantized components and auxiliary storage. These measurements characterize the tested configuration; additional radius storage can still matter when available memory is close to a deployment limit.

\begin{table}[!htbp]
  \centering
  \caption{Measured GPU-memory consumption of quantized Qwen3-VL-8B-Instruct. Payload refers to encoded weights and does not include auxiliary storage.}
  \label{tab:gpu-memory}
  \small
  \setlength{\tabcolsep}{12pt}
  \begin{tabular}{lcc}
    \toprule
    Method & Payload (bits/weight) & GPU memory (GB) $\downarrow$ \\
    \midrule
    QTIP & 2.00 & 5.02 \\
    ParoQuant & 2.00 & 5.24 \\
    SPHQuant & 2.25 & 5.35 \\
    \bottomrule
  \end{tabular}
\end{table}

\subsection{Comparison at Matched Encoded Bit-widths}
\label{app:matched-bits}
To separate accuracy gains from additional weight-storage capacity, we compare SPHQuant, RTN, and ParoQuant on Qwen3-VL-8B-Instruct at 3.00 and 3.125 encoded bits per weight. Tab.~\ref{tab:matched-bits} shows that SPHQuant obtains the highest MMMU and TextVQA accuracy among these methods at both budgets. At 3.00 bits, the margins over ParoQuant are 0.1 and 0.3 percentage points; at 3.125 bits, they increase to 0.7 and 0.8 percentage points. These comparisons complement the nominal W2/W3 results by making the encoded weight budget explicit.

\begin{table}[!htbp]
  \centering
  \caption{Accuracy at matched encoded weight budgets on Qwen3-VL-8B-Instruct. The bit-widths refer to encoded weights, excluding auxiliary storage.}
  \label{tab:matched-bits}
  \small
  \setlength{\tabcolsep}{9pt}
  \begin{tabular}{clcc}
    \toprule
    Bits/weight & Method & MMMU\_val $\uparrow$ & TextVQA $\uparrow$ \\
    \midrule
    3.00 & RTN & 46.9 & 75.4 \\
    3.00 & ParoQuant & 50.5 & 79.5 \\
    3.00 & SPHQuant & \textbf{50.6} & \textbf{79.8} \\
    \midrule
    3.125 & RTN & 47.8 & 76.3 \\
    3.125 & ParoQuant & 51.0 & 80.2 \\
    3.125 & SPHQuant & \textbf{51.7} & \textbf{81.0} \\
    \bottomrule
  \end{tabular}
\end{table}

\section{Limitations}
\label{app:limitations}
SPHQuant is designed for efficient extreme low-bit weight-only quantization of VLMs, but several limitations remain. First, our throughput evaluation covers RTX A6000 and RTX 4090, but does not establish performance on embedded or mobile-class devices. The added RTX 4090 evaluation uses batch size 1 and a fixed sequence-length configuration; broader batch-size and context-length studies remain necessary. Since the practical benefit of low-bit inference depends on memory hierarchy, shared-memory capacity, instruction scheduling, and kernel integration, the efficiency of SPHQuant should be further validated on a wider range of devices. The additional radius storage can also affect whether a model fits when memory capacity is close to its footprint. Second, although we evaluate representative multimodal benchmarks, VLMs are often deployed in diverse real-world applications such as mobile OCR, document understanding, visual assistance, and embodied reasoning. More task-specific evaluations under realistic latency, input-resolution, and generation-length constraints are needed to fully characterize the accuracy-efficiency trade-off in deployment.

\section{Broader Impacts}
\label{app:broader-impacts}
By reducing the memory footprint and improving decoding throughput of VLMs, SPHQuant may help make multimodal models more accessible on resource-constrained devices. This can support privacy-preserving on-device inference, reduce dependence on cloud connectivity, and lower the energy and hardware cost of deploying useful visual-language applications.


\begin{thebibliography}{33}
\providecommand{\natexlab}[1]{#1}
\providecommand{\url}[1]{\texttt{#1}}
\expandafter\ifx\csname urlstyle\endcsname\relax
  \providecommand{\doi}[1]{doi: #1}\else
  \providecommand{\doi}{doi: \begingroup \urlstyle{rm}\Url}\fi

\bibitem[Bai et~al.(2025)Bai, Cai, Chen, Chen, Chen, Cheng, Deng, Ding, Gao,
  Ge, Ge, Guo, Huang, Huang, Huang, Hui, Jiang, Li, Li, Li, Li, Lin, Lin, Liu,
  Liu, Liu, Liu, Liu, Liu, Lu, Luo, Lv, Men, Meng, Ren, Ren, Song, Sun, Tang,
  Tu, Wan, Wang, Wang, Wang, Wang, Xie, Xu, Xu, Xu, Yang, Yang, Yang, Yang, Yu,
  Zhang, Zhang, Zhang, Zheng, Zhong, Zhou, Zhou, Zhou, Zhu, and
  Zhu]{bai2025qwen3vltechnicalreport}
Shuai Bai, Yuxuan Cai, Ruizhe Chen, Keqin Chen, Xionghui Chen, Zesen Cheng,
  Lianghao Deng, Wei Ding, Chang Gao, Chunjiang Ge, Wenbin Ge, Zhifang Guo,
  Qidong Huang, Jie Huang, Fei Huang, Binyuan Hui, Shutong Jiang, Zhaohai Li,
  Mingsheng Li, Mei Li, Kaixin Li, Zicheng Lin, Junyang Lin, Xuejing Liu,
  Jiawei Liu, Chenglong Liu, Yang Liu, Dayiheng Liu, Shixuan Liu, Dunjie Lu,
  Ruilin Luo, Chenxu Lv, Rui Men, Lingchen Meng, Xuancheng Ren, Xingzhang Ren,
  Sibo Song, Yuchong Sun, Jun Tang, Jianhong Tu, Jianqiang Wan, Peng Wang,
  Pengfei Wang, Qiuyue Wang, Yuxuan Wang, Tianbao Xie, Yiheng Xu, Haiyang Xu,
  Jin Xu, Zhibo Yang, Mingkun Yang, Jianxin Yang, An~Yang, Bowen Yu, Fei Zhang,
  Hang Zhang, Xi~Zhang, Bo~Zheng, Humen Zhong, Jingren Zhou, Fan Zhou, Jing
  Zhou, Yuanzhi Zhu, and Ke~Zhu.
\newblock {Qwen3-VL Technical Report}, 2025.
\newblock URL \url{https://arxiv.org/abs/2511.21631}.

\bibitem[Chai et~al.(2026)Chai, Chen, Zhu, Li, Guo, and
  Zhang]{chai2026quantvsr}
Bowen Chai, Zheng Chen, Libo Zhu, Wenbo Li, Yong Guo, and Yulun Zhang.
\newblock {{QuantVSR}: Low-Bit Post-Training Quantization for Real-World Video
  Super-Resolution}.
\newblock \emph{Proceedings of the AAAI Conference on Artificial Intelligence},
  40\penalty0 (4):\penalty0 2689--2697, 2026.
\newblock \doi{10.1609/aaai.v40i4.37257}.
\newblock URL \url{https://ojs.aaai.org/index.php/AAAI/article/view/37257}.

\bibitem[Chee et~al.(2024)Chee, Cai, Kuleshov, and
  Sa]{chee2024quip2bitquantizationlarge}
Jerry Chee, Yaohui Cai, Volodymyr Kuleshov, and Christopher~De Sa.
\newblock {QuIP: 2-Bit Quantization of Large Language Models With Guarantees},
  2024.
\newblock URL \url{https://arxiv.org/abs/2307.13304}.

\bibitem[Chen et~al.(2024)Chen, Li, Dong, Zhang, Zang, Chen, Duan, Wang, Qiao,
  Lin, and Zhao]{chen2024rightwayevaluatinglarge}
Lin Chen, Jinsong Li, Xiaoyi Dong, Pan Zhang, Yuhang Zang, Zehui Chen, Haodong
  Duan, Jiaqi Wang, Yu~Qiao, Dahua Lin, and Feng Zhao.
\newblock {Are We on the Right Way for Evaluating Large Vision-Language
  Models?}, 2024.
\newblock URL \url{https://arxiv.org/abs/2403.20330}.

\bibitem[Chen et~al.(2025)Chen, Zhang, Liu, Zhang, Wang, Fu, and
  Zhang]{chen2025quantdemoire}
Zheng Chen, Kewei Zhang, Xiaoyang Liu, Weihang Zhang, Mengfan Wang, Yifan Fu,
  and Yulun Zhang.
\newblock {{QuantDemoire}: Quantization with Outlier Aware for Image
  Demoir\'eing}.
\newblock \emph{arXiv preprint arXiv:2510.04066}, 2025.
\newblock URL \url{https://arxiv.org/abs/2510.04066}.

\bibitem[Chu et~al.(2023)Chu, Qiao, Lin, Xu, Yang, Hu, Wei, Zhang, Zhang, Wei,
  and Shen]{chu2023mobilevlm}
Xiangxiang Chu, Limeng Qiao, Xinyang Lin, Shuang Xu, Yang Yang, Yiming Hu, Fei
  Wei, Xinyu Zhang, Bo~Zhang, Xiaolin Wei, and Chunhua Shen.
\newblock {{MobileVLM}: A Fast, Strong and Open Vision Language Assistant for
  Mobile Devices}, 2023.
\newblock URL \url{https://arxiv.org/abs/2312.16886}.

\bibitem[Chu et~al.(2024)Chu, Qiao, Zhang, Xu, Wei, Yang, Sun, Hu, Lin, Zhang,
  and Shen]{chu2024mobilevlmv2}
Xiangxiang Chu, Limeng Qiao, Xinyu Zhang, Shuang Xu, Fei Wei, Yang Yang,
  Xiaofei Sun, Yiming Hu, Xinyang Lin, Bo~Zhang, and Chunhua Shen.
\newblock {{MobileVLM V2}: Faster and Stronger Baseline for Vision Language
  Model}, 2024.
\newblock URL \url{https://arxiv.org/abs/2402.03766}.

\bibitem[Farabet and Lacombe(2026)]{farabet2026gemma4}
Clement Farabet and Olivier Lacombe.
\newblock {{Gemma 4}: Byte for byte, the most capable open models}, April 2026.
\newblock URL
  \url{https://blog.google/innovation-and-ai/technology/developers-tools/gemma-4/}.

\bibitem[Frantar et~al.(2022)Frantar, Ashkboos, Hoefler, and
  Alistarh]{frantar-gptq}
Elias Frantar, Saleh Ashkboos, Torsten Hoefler, and Dan Alistarh.
\newblock {{GPTQ}: Accurate Post-training Compression for Generative Pretrained
  Transformers}.
\newblock \emph{arXiv preprint arXiv:2210.17323}, 2022.

\bibitem[{Google DeepMind}(2026)]{google2026gemma4modelcard}
{Google DeepMind}.
\newblock {{Gemma 4} Model Card}, 2026.
\newblock URL \url{https://ai.google.dev/gemma/docs/core/model_card_4}.
\newblock Last updated 2026-04-17.

\bibitem[Jin et~al.(2025)Jin, Li, Liu, Gu, Wu, Jiang, He, Zhao, Tan, Gan, Wang,
  Wang, and Ma]{jin2025efficientmllmsurvey}
Yizhang Jin, Jian Li, Yexin Liu, Tianjun Gu, Kai Wu, Zhengkai Jiang, Muyang He,
  Bo~Zhao, Xin Tan, Zhenye Gan, Yabiao Wang, Chengjie Wang, and Lizhuang Ma.
\newblock {Efficient Multimodal Large Language Models: A Survey}.
\newblock \emph{Visual Intelligence}, 3\penalty0 (27), 2025.
\newblock \doi{10.1007/s44267-025-00099-6}.
\newblock URL \url{https://arxiv.org/abs/2405.10739}.

\bibitem[Kingma and Ba(2017)]{kingma2017adammethodstochasticoptimization}
Diederik~P. Kingma and Jimmy Ba.
\newblock {Adam: A Method for Stochastic Optimization}, 2017.
\newblock URL \url{https://arxiv.org/abs/1412.6980}.

\bibitem[Li et~al.(2024)Li, Hu, Ning, Liu, Hong, Jia, Li, Yan, Ran, Dai, Yan,
  Yang, and Wang]{li2024mbq}
Shiyao Li, Yingchun Hu, Xuefei Ning, Xihui Liu, Ke~Hong, Xiaotao Jia, Xiuhong
  Li, Yaqi Yan, Pei Ran, Guohao Dai, Shengen Yan, Huazhong Yang, and Yu~Wang.
\newblock {MBQ: Modality-Balanced Quantization for Large Vision-Language
  Models}, 2024.
\newblock URL \url{https://arxiv.org/abs/2412.19509}.

\bibitem[Liang et~al.(2026)Liang, Chen, Zhang, Han, and
  Liu]{liang2026paroquant}
Yesheng Liang, Haisheng Chen, Zihan Zhang, Song Han, and Zhijian Liu.
\newblock {{ParoQuant: Pairwise Rotation Quantization for Efficient Reasoning
  LLM Inference}}.
\newblock In \emph{International Conference on Learning Representations
  (ICLR)}, 2026.

\bibitem[Lin et~al.(2024)Lin, Tang, Tang, Yang, Chen, Wang, Xiao, Dang, Gan,
  and Han]{lin2023awq}
Ji~Lin, Jiaming Tang, Haotian Tang, Shang Yang, Wei-Ming Chen, Wei-Chen Wang,
  Guangxuan Xiao, Xingyu Dang, Chuang Gan, and Song Han.
\newblock {AWQ: Activation-aware Weight Quantization for LLM Compression and
  Acceleration}.
\newblock In \emph{MLSys}, 2024.

\bibitem[Lin et~al.(2015)Lin, Maire, Belongie, Bourdev, Girshick, Hays, Perona,
  Ramanan, Zitnick, and Dollár]{lin2015microsoftcococommonobjects}
Tsung-Yi Lin, Michael Maire, Serge Belongie, Lubomir Bourdev, Ross Girshick,
  James Hays, Pietro Perona, Deva Ramanan, C.~Lawrence Zitnick, and Piotr
  Dollár.
\newblock {Microsoft COCO: Common Objects in Context}, 2015.
\newblock URL \url{https://arxiv.org/abs/1405.0312}.

\bibitem[Lu et~al.(2022)Lu, Mishra, Xia, Qiu, Chang, Zhu, Tafjord, Clark, and
  Kalyan]{lu2022learnexplainmultimodalreasoning}
Pan Lu, Swaroop Mishra, Tony Xia, Liang Qiu, Kai-Wei Chang, Song-Chun Zhu,
  Oyvind Tafjord, Peter Clark, and Ashwin Kalyan.
\newblock {Learn to Explain: Multimodal Reasoning via Thought Chains for
  Science Question Answering}, 2022.
\newblock URL \url{https://arxiv.org/abs/2209.09513}.

\bibitem[{NVIDIA Corporation}(2026{\natexlab{a}})]{nvidia_ada_tuning_guide}
{NVIDIA Corporation}.
\newblock {NVIDIA Ada GPU Architecture Tuning Guide}, 2026{\natexlab{a}}.
\newblock URL \url{https://docs.nvidia.com/cuda/ada-tuning-guide/index.html}.

\bibitem[{NVIDIA
  Corporation}(2026{\natexlab{b}})]{nvidia_cuda_programming_guide}
{NVIDIA Corporation}.
\newblock {CUDA C++ Programming Guide}, 2026{\natexlab{b}}.
\newblock URL \url{https://docs.nvidia.com/cuda/cuda-c-programming-guide/}.

\bibitem[Qin et~al.(2026)Qin, Li, Chen, Zhang, Kong, and Zhang]{qin2026veq}
Guangshuo Qin, Zhiteng Li, Zheng Chen, Weihang Zhang, Linghe Kong, and Yulun
  Zhang.
\newblock {VEQ: Modality-Adaptive Quantization for MoE Vision-Language Models}.
\newblock \emph{arXiv preprint arxiv:2602.01037}, 2026.

\bibitem[{Qwen Team}(2026{\natexlab{a}})]{qwen3.5}
{Qwen Team}.
\newblock {{Qwen3.5}: Towards Native Multimodal Agents}, February
  2026{\natexlab{a}}.
\newblock URL \url{https://qwen.ai/blog?id=qwen3.5}.

\bibitem[{Qwen Team}(2026{\natexlab{b}})]{qwen36_35b_a3b}
{Qwen Team}.
\newblock {{Qwen3.6-35B-A3B}: Agentic Coding Power, Now Open to All}, April
  2026{\natexlab{b}}.
\newblock URL \url{https://qwen.ai/blog?id=qwen3.6-35b-a3b}.

\bibitem[Singh et~al.(2019)Singh, Natarajan, Shah, Jiang, Chen, Batra, Parikh,
  and Rohrbach]{singh2019vqamodelsread}
Amanpreet Singh, Vivek Natarajan, Meet Shah, Yu~Jiang, Xinlei Chen, Dhruv
  Batra, Devi Parikh, and Marcus Rohrbach.
\newblock {Towards VQA Models That Can Read}, 2019.
\newblock URL \url{https://arxiv.org/abs/1904.08920}.

\bibitem[Tseng et~al.(2024{\natexlab{a}})Tseng, Chee, Sun, Kuleshov, and
  Sa]{tseng2024quipbetterllmquantization}
Albert Tseng, Jerry Chee, Qingyao Sun, Volodymyr Kuleshov, and Christopher~De
  Sa.
\newblock {QuIP\#: Even Better LLM Quantization with Hadamard Incoherence and
  Lattice Codebooks}, 2024{\natexlab{a}}.
\newblock URL \url{https://arxiv.org/abs/2402.04396}.

\bibitem[Tseng et~al.(2024{\natexlab{b}})Tseng, Sun, Hou, and
  Sa]{tseng2024qtip}
Albert Tseng, Qingyao Sun, David Hou, and Christopher~De Sa.
\newblock {{QTIP}: Quantization with Trellises and Incoherence Processing}.
\newblock In \emph{The Thirty-eighth Annual Conference on Neural Information
  Processing Systems}, 2024{\natexlab{b}}.
\newblock URL \url{https://openreview.net/forum?id=7sdkLVuYCU}.

\bibitem[van~der Ouderaa et~al.(2024)van~der Ouderaa, Croci, Hilmkil, and
  Hensman]{vanderouderaa2024pyramid}
Tycho F.~A. van~der Ouderaa, Maximilian~L. Croci, Agrin Hilmkil, and James
  Hensman.
\newblock {Pyramid Vector Quantization for LLMs}, 2024.
\newblock URL \url{https://arxiv.org/abs/2410.16926}.

\bibitem[Wang et~al.(2025)Wang, Gao, Gu, Pu, Cui, Wei, Liu, Jing, Ye, Shao,
  et~al.]{wang2025internvl3_5}
Weiyun Wang, Zhangwei Gao, Lixin Gu, Hengjun Pu, Long Cui, Xingguang Wei,
  Zhaoyang Liu, Linglin Jing, Shenglong Ye, Jie Shao, et~al.
\newblock {{InternVL3.5}: Advancing Open-Source Multimodal Models in
  Versatility, Reasoning, and Efficiency}.
\newblock \emph{arXiv preprint arXiv:2508.18265}, 2025.
\newblock URL \url{https://arxiv.org/abs/2508.18265}.

\bibitem[Wolf et~al.(2020)Wolf, Debut, Sanh, Chaumond, Delangue, Moi, Cistac,
  Rault, Louf, Funtowicz, Davison, Shleifer, von Platen, Ma, Jernite, Plu, Xu,
  Le~Scao, Gugger, Drame, Lhoest, and Rush]{wolf-etal-2020-transformers}
Thomas Wolf, Lysandre Debut, Victor Sanh, Julien Chaumond, Clement Delangue,
  Anthony Moi, Pierric Cistac, Tim Rault, Remi Louf, Morgan Funtowicz, Joe
  Davison, Sam Shleifer, Patrick von Platen, Clara Ma, Yacine Jernite, Julien
  Plu, Canwen Xu, Teven Le~Scao, Sylvain Gugger, Mariama Drame, Quentin Lhoest,
  and Alexander Rush.
\newblock {Transformers: State-of-the-Art Natural Language Processing}.
\newblock In Qun Liu and David Schlangen, editors, \emph{Proceedings of the
  2020 Conference on Empirical Methods in Natural Language Processing: System
  Demonstrations}, pages 38--45, Online, October 2020. Association for
  Computational Linguistics.
\newblock \doi{10.18653/v1/2020.emnlp-demos.6}.
\newblock URL \url{https://aclanthology.org/2020.emnlp-demos.6/}.

\bibitem[Wu et~al.(2026)Wu, Chen, Xu, Chai, Guo, Liu, Kong, and
  Zhang]{wu2026lsgquant}
Tianxing Wu, Zheng Chen, Cirou Xu, Bowen Chai, Yong Guo, Yutong Liu, Linghe
  Kong, and Yulun Zhang.
\newblock {{LSGQuant}: Layer-Sensitivity Guided Quantization for One-Step
  Diffusion Real-World Video Super-Resolution}.
\newblock \emph{arXiv preprint arXiv:2602.03182}, 2026.
\newblock URL \url{https://arxiv.org/abs/2602.03182}.

\bibitem[{xAI Team}(2024)]{xai2024realworldqa}
{xAI Team}.
\newblock {RealWorldQA: A Real-World Multimodal Question Answering Benchmark},
  2024.
\newblock URL \url{https://huggingface.co/datasets/xai-org/RealworldQA}.

\bibitem[Yue et~al.(2024)Yue, Ni, Zhang, Zheng, Liu, Zhang, Stevens, Jiang,
  Ren, Sun, Wei, Yu, Yuan, Sun, Yin, Zheng, Yang, Liu, Huang, Sun, Su, and
  Chen]{yue2024mmmumassivemultidisciplinemultimodal}
Xiang Yue, Yuansheng Ni, Kai Zhang, Tianyu Zheng, Ruoqi Liu, Ge~Zhang, Samuel
  Stevens, Dongfu Jiang, Weiming Ren, Yuxuan Sun, Cong Wei, Botao Yu, Ruibin
  Yuan, Renliang Sun, Ming Yin, Boyuan Zheng, Zhenzhu Yang, Yibo Liu, Wenhao
  Huang, Huan Sun, Yu~Su, and Wenhu Chen.
\newblock {MMMU: A Massive Multi-discipline Multimodal Understanding and
  Reasoning Benchmark for Expert AGI}, 2024.
\newblock URL \url{https://arxiv.org/abs/2311.16502}.

\bibitem[Yue et~al.(2025{\natexlab{a}})Yue, Zheng, Ni, Wang, Zhang, Tong, Sun,
  Yu, Zhang, Sun, Su, Chen, and
  Neubig]{yue2025mmmuprorobustmultidisciplinemultimodal}
Xiang Yue, Tianyu Zheng, Yuansheng Ni, Yubo Wang, Kai Zhang, Shengbang Tong,
  Yuxuan Sun, Botao Yu, Ge~Zhang, Huan Sun, Yu~Su, Wenhu Chen, and Graham
  Neubig.
\newblock {MMMU-Pro: A More Robust Multi-discipline Multimodal Understanding
  Benchmark}, 2025{\natexlab{a}}.
\newblock URL \url{https://arxiv.org/abs/2409.02813}.

\bibitem[Yue et~al.(2025{\natexlab{b}})Yue, Xu, Yuan, Yang, Wu, and
  Nie]{yue2025pcdvqenhancingvectorquantization}
Yuxuan Yue, Zukang Xu, Zhihang Yuan, Dawei Yang, Jianlong Wu, and Liqiang Nie.
\newblock {PCDVQ: Enhancing Vector Quantization for Large Language Models via
  Polar Coordinate Decoupling}, 2025{\natexlab{b}}.
\newblock URL \url{https://arxiv.org/abs/2506.05432}.

\end{thebibliography}
\end{document}